\documentclass[conference]{IEEEtran}

\newif\ifanon
\anonfalse

\usepackage[utf8]{inputenc}
\usepackage[T1]{fontenc}
\usepackage[hidelinks]{hyperref}
\usepackage{url}
\usepackage{booktabs}
\usepackage{amsmath}
\usepackage{amssymb}
\usepackage{amsfonts}
\usepackage{microtype}
\usepackage{graphicx}
\usepackage{pgfplots}
\usepackage{subcaption}
\pgfplotsset{compat=1.18}
\usepgfplotslibrary{groupplots}
\definecolor{okblue}{rgb}{0,0.447,0.698}
\definecolor{okverm}{rgb}{0.835,0.369,0}
\definecolor{okgreen}{rgb}{0,0.620,0.451}

\begin{document}

\title{Making Agents More Consistent:\\Skills Should Form Habits for Repeat Tasks}

\ifanon
  \author{\IEEEauthorblockN{Anonymous Author(s)}
  \IEEEauthorblockA{Submission under double-blind review}}
\else
  \author{\IEEEauthorblockN{Travis Weber and Rohit Taneja}
  \IEEEauthorblockA{Pheo Inc\\
  \{travis,rohit\}@pheo.ai}}
\fi

\maketitle

\begin{abstract}
On repeated work, agents are inconsistent. We ran 42 tasks three times each and found that, depending on the model, $38\%$ to $74\%$ of them returned answers that did not agree. Consistency is the property a buyer, an auditor, or a regulator requires, and agents do not have it. They are wasteful too: $95.3\%$ to $97.2\%$ of what an agent generates goes to re-deriving a plan the system already knows.

We propose skill habit formation to improve the consistency of agent output while reducing inference cost. To form a habit, an agent mines its own execution history for candidate skills, faster and deterministic variants that compete against the incumbent instead of replacing it. A candidate declares the region of input space it claims, so the common case runs as a script and the rest falls through to reasoning. Every ancestor stays executable, which makes reversion a pointer operation. Four gates of ascending cost admit candidates, the central one testing a candidate's execution trace against a retained reference within a tolerance measured from that reference's own run-to-run variability.

Measured on text-to-SQL, three of four reasoning arms reproduced their own output on 11 to 13 of 42 repeated questions and the fourth on 26 of 42, while a habit-formed variant reproduced on all 456 dispatches we repeated and was non-inferior to every arm it replaced ($p<0.0001$). Replay is exact on the slice it serves, so the work becomes auditable. Habit-formed skills cost less as well, using 14\% to 56\% fewer tokens per served request and turning net positive after 7 to 53 reuses, though the cost case is conditional in ways the consistency case is not.

We also measured what habit formation costs in accuracy. The guard admitted work it should have deferred on 2.6\% of natural paraphrases and 26\% of inputs near its boundary, and 11 of 13 such failures were invisible to the trace-conformance gate at any threshold. Deterministic errors repeat exactly: a bad habit is as reliable as a good one, and that is the price of the property that makes the system auditable. Separating the routing decision from parameter extraction raised end-to-end accuracy from 0.888 to 0.952 at 43\% of the cost and localized the errors that remain, while keeping outputs deterministic. We state plainly the failure modes of habit formation and what remains unmeasured.
\end{abstract}


\section{Introduction}
\label{sec:intro}

Agentic systems have reached a level of capability at which the limiting factor in deployment is now output consistency, not task range. A skill that produces excellent work on one occasion and materially different work on the next cannot be delegated with confidence, audited after the fact, or sold into any setting where a person must answer for the result. The four reasoning arms we measure in Section~\ref{sec:evidence} disagree with themselves on $38\%$ to $74\%$ of repeated tasks, so this is a present property of capable systems and not a worry about future ones.

These systems are also more expensive to operate than they need to be, in a way that is specific and measurable. Most of what a reasoning agent generates goes into \emph{planning}: establishing what sequence of steps a task requires before any step is carried out. On the same workload that share is $95.3\%$ to $97.2\%$ of generated tokens across the three arms that meter reasoning separately, and $62.7\%$ to $75.0\%$ of all tokens billed, on a task whose plan never changes. Much of that planning is \emph{rediscovery}. An agent derives a workflow, completes the task, discards the derivation, and arrives at the same workflow again when a similar request appears. Where the same workflow has been derived four times in succession, the fourth derivation has purchased nothing the first did not already provide, and the procedure would be better held as a script the agent calls. This is the classical target of macro-operator learning and case-based planning, an old problem rather than a new one about agents specifically, and the failure mode that follows it, the \emph{utility problem} \cite{minton1990}, is a known negative result about this program: cached knowledge slows a system down unless it is selectively retained. Section~\ref{sec:economy} treats it as a live constraint, not a footnote.

Unreliability and rediscovery share a cause. An agent with no durable record of how it performed a task must reconstruct the approach on each occasion, and that reconstruction is both the source of duplicated cost and the source of variation. Existing responses address neither: evaluation harnesses score outputs after the fact and stay silent on how those outputs were produced; prompt and pipeline optimizers improve an artifact in place, leaving no way to ask what the improvement replaced; guardrails constrain the envelope of behavior while saying nothing about the route taken inside it. In each case the system forgoes the cheapest available source of assurance: comparison against what it used to do.

We propose \textbf{agentic skill habit formation}. A skill is treated as a \emph{lineage} of executable variants implementing one declared capability contract. By varying its own execution history through replay, perturbation, and short simulated excursions, the agent proposes \emph{candidate skills}. Candidates compete against the incumbent and never overwrite it. Two structural commitments carry the argument. \emph{Variants are partial}: a variant declares a \emph{coverage predicate} naming the region of input space it claims, so habit formation requires only that a fast variant match reasoning within a region it can recognize. That is a far weaker requirement than matching it everywhere, and it is the discipline a polymorphic inline cache observes \cite{holzle1991}. \emph{Ancestors stay warm}: because the reasoning path continues to serve the tail, production exercises it daily, so reversion is a pointer operation instead of an act of faith.

\emph{The name is an analogy, and we use it as one.} In behavioral psychology a habit is a response that has acquired \emph{automaticity}: cue-triggered, efficient, and insensitive to the value of its outcome. The split between controlled and automatic processing is the one this architecture hard-wires, with the reasoning path on one side, the scripted variant on the other, and the coverage guard deciding which runs. What we borrow is that structure and not a mechanism: a procedure that was once derived becomes triggered by recognition of its cue, and what we have to guard against is the insensitivity that makes it cheap. We claim no resemblance between the underlying processes, and where we borrow an actual experimental technique rather than a word, outcome devaluation in \S\ref{sec:recovery}, we say so at the point of use. A reader who prefers to read ``habit formation'' throughout as ``coverage-gated specialization of a retained skill'' will lose nothing.

We cannot prove that a fast variant means the same thing as the reasoning it replaces, since no such proof is available for a stochastic model with no formal semantics. The situation is nonetheless familiar: testing against an independently produced implementation where no specification exists is Weyuker's \emph{pseudo-oracle} \cite{weyuker1982}. We adopt it, with well-understood limits.

Prior work governs \emph{whether} an agent may act autonomously, through graduated per-skill autonomy and a correction-rate graduation condition \cite{weber2026}. That line of work leaves open how a skill is internally composed. This paper supplies the missing half: what verifies a candidate before it is trusted with traffic.

\textbf{Contributions.} (1) A lineage architecture in which candidate skills are admitted through four derived, ascending-cost gates, not authored per-candidate tests. (2) A trace-conformance gate that measures its own discriminative power from the reference's run-to-run variability, requiring no declared equivalence relation. (3) Preliminary evidence (\S\ref{sec:evidence}) on two workloads, led by exact replay. A habit-formed variant reproduces its own output on everything it serves, where the arms it replaces manage between a quarter and two-thirds. Token reduction and a break-even reuse count follow as the secondary and more conditional claim, together with a controlled follow-up locating where coverage-gated dispatch breaks (phrasing, rather than task complexity) and a guard redesign that repairs it; and a retrieval-routing replication across three router families, two of them at three seeds. (4) Two failure modes that bear on any admission-gated learner: a non-inferiority gate that can deadlock the learner it filters, and mined corrections that are one-sided by construction and therefore cannot alone train a guard with no prior. (5) A falsifiable measurement plan (\S\ref{sec:limits}) for the claims still outstanding.

\section{Related work}
\label{sec:related}

\emph{Dynamic compilation.} A tracing JIT interprets a program, compiles hot paths, installs guards on the facts a compiled trace assumed, and deoptimizes when a guard fails \cite{bala2000,gal2009,bolz2009,holzle1992}. We borrow the discipline that every fast path carries a guard with a defined path back, and reject one assumption: a JIT can discard compiled code because the interpreter is deterministic and, in verified systems, provably equivalent to it \cite{barriere2021}. None of that transfers to a stochastic reasoner, so we keep the slow path running rather than merely recoverable.

\emph{Testing without a specification.} Differential and metamorphic techniques compare a program against an independent implementation or against relations its outputs must satisfy \cite{weyuker1982}. Our trace-conformance gate (\S\ref{sec:trace}) is a differential test whose pseudo-oracle is the system's own retained ancestor, over execution traces rather than outputs.

\emph{Agent self-improvement.} Pipeline compilers optimize declarative LM programs against a metric \cite{khattab2024}; library learning discovers reusable subroutines by search \cite{ellis2021}; workflow-memory methods induce reusable routines from trajectories and report large step-count and success-rate gains \cite{wang2025awm}; skill libraries built from executable code give large sample-efficiency gains in open-ended agents \cite{wang2023voyager}. All improve an artifact in place or grow a library monotonically; none retains a competing lineage or answers to a governance principal that can revoke a promotion.

\emph{Specialization, deferral, and reliability measurement.} Binding-time analysis \cite{jones1993} supplies the static/dynamic distinction our coverage predicates need; learning with rejection \cite{cortes2016,chow1970} supplies a cheap predictor that defers when unsure. pass$^{k}$, the probability that $k$ independent attempts on a task all succeed, shows state-of-the-art tool-using agents to be ``terribly inconsistent'' under repetition \cite{yao2024taubench}, independent evidence for the unreliability half of our motivation. He et al.\ \cite{he2025nondeterminism} localize temperature-0 run-to-run variation to batch-dependent kernel behavior rather than non-associativity, so a scripted variant's determinism is structural (no sampling occurs), not a tuning artifact of the baseline it replaces. Cuadron et al.\ \cite{cuadron2025overthinking} show that selecting the lowest-overthinking solution on SWE-bench Verified improves accuracy \emph{and} cuts cost, direct evidence that planning-token duplication is real and often unproductive. Cascade and routing systems report up to 98\% cost reduction from deferring only hard cases to an expensive path \cite{chen2023frugalgpt}, and compiler-style parallel tool dispatch reports large joint latency/cost reductions \cite{kim2024llmcompiler}; we adopt the coverage-gated dispatch discipline these depend on, applied to whole procedures rather than single calls.

\emph{Convergent industry practice.} Three widely used coding-agent products shipped persistent, reusable procedures within roughly nine months of each other. What their public documentation describes, in each case, is a skill as a single mutable artifact updated in place, rather than a lineage of variants. That is the distinction \S\ref{sec:intro} turns on.

\section{The lineage model}
\label{sec:lineage}

A \emph{skill} is a declared capability contract together with an ordered set of variants implementing it. Each variant is placed by how much of its execution is model inference, from a fully reasoned bundle of instructions down to a deterministic script. One variant, the \textbf{anchor}, is structurally privileged and may never be pruned: the slowest retained ancestor, the terminal fallthrough, and the reference against which all conformance is measured.

\textbf{Coverage-gated dispatch.} Each variant $v$ carries a coverage predicate $\chi_v$ and a \emph{coverage guard} that evaluates it. Dispatch walks the lineage from the lowest rung upward; the first variant whose guard admits the input serves it, and the anchor admits everything. The guard is asymmetric. A false negative sends a head case to reasoning and costs tokens. A false positive sends a tail case to a script written for a different situation, producing confident wrong work with no reasoning step in which the mismatch could surface. The guard is therefore tuned for precision over recall, via a rejector trained under a learning-with-rejection objective \cite{cortes2016,chow1970}. Coverage makes the traffic-weighted share of executions served without reasoning directly measurable, and, read as an assurance figure, it reports the fraction of work whose procedure can be exhibited rather than only scored.

\textbf{The determinism tension.} A deterministic variant is simultaneously more auditable and more brittle: it cannot notice the world has moved, and fails quietly where a reasoning path fails loudly. Partial coverage bounds the resulting exposure directly: a variant claiming eighty percent of traffic still leaves twenty percent in continuous contact with whatever is novel.

\subsection{The lineage economy}
\label{sec:economy}

Unlimited retention degrades into hoarding. SOAR and ACT-R both lost performance as learned rules accumulated, and the remedy in both cases was forgetting \cite{minton1990,kennedy2003}; a lineage that only grows re-enacts the utility problem with better provenance. We therefore generate candidates in mutated batches over two cheap search spaces, the coverage predicate and the guard parameters, while keeping retention selective: the anchor and any variant currently serving traffic are unprunable; everything else is pruned by non-selection over a window; and pruning is recorded so a gap in the lineage remains visible. Batch generation carries a statistical cost of its own: testing many candidates against one key set inflates false admissions, so admission decisions within a batch control the false discovery rate across the batch \cite{benjamini1995} rather than applying a per-candidate threshold.

\section{Candidate generation}
\label{sec:generation}

Candidates come from a forge step that mines the trace spine for stable structure. Three sources feed it, in descending order of trustworthiness: \emph{replay} of historical traces, where per-decision-point conditional entropy freezes near-deterministic junctures, distills high-entropy-but-learnable ones, and excludes the rest from the coverage predicate entirely; \emph{counterfactual and perturbed replay}, the cheapest probe of coverage-boundary sensitivity; and \emph{bounded simulation} \cite{sutton1990}, kept short and always branched from a real execution state, because learned-model error compounds over the rollout horizon \cite{janner2019} and recursive training on synthetic output is where distributional tails are lost \cite{shumailov2024}.

A second candidate type, \textbf{proposed guardrails}, is mined from what corrections teach: a correction chiefly marks a \emph{boundary}, and a cluster of corrections in one region induces a deterministic assertion that would have blocked the corrected outputs. The proposal is learned while the guardrail itself stays deterministic and binary, entering the tenant set only by the same human authorization a promotion requires.

\section{Admission: four derived gates}
\label{sec:gates}

Unattended habit formation is viable because admission is cheap to \emph{construct}, not because it is strict. Authored input is consumed once per capability or tenant and never per candidate: the guardrail set, the reversibility tier, the non-inferiority margin, and the dependency pinning record. Authoring cost is therefore $O(\text{capabilities})$ while candidate volume is what scales. That ratio is why generating candidates in bulk stays affordable.

\begin{table}[t]
\caption{The four admission gates, ordered by resource cost.}
\label{tab:gates}
\footnotesize
\centering
\begin{tabular}{@{}llll@{}}
\toprule
 & \textbf{Derived from} & \textbf{Resource} & \textbf{Catches} \\
\midrule
C0 & lineage manifest & probe & no way back \\
C1 & inherited guardrails & static & a forbidden action \\
C2 & key traces, spine & replay & the wrong route \\
C3 & rubric, $H_{\text{auth}}$ & traffic, human & conforms but worse \\
\bottomrule
\end{tabular}
\end{table}

\textbf{C0: recoverability.} An escape route never checked provides no assurance. Before a candidate is considered, the variant that would receive its traffic on withdrawal must pass an \emph{executability probe} confirming it still binds to its pinned model, prompt, tool schemas, and retrieval surface. Probe failure blocks promotion, leaves execution untouched, and raises a repair task; where re-pinning does not suffice, \emph{re-anchoring} onto a current model is a demotion event, since it invalidates the C2 conformance apparatus built against the prior anchor.

\textbf{C1: governance conformance.} A \emph{guardrail} is a deterministic, binary, blocking assertion; an assertion requiring inference is a scorer, not a guardrail. C1 is a static subset check of a candidate's declared guardrails against its parent's: decidable, free, and unfalsifiable by the candidate's own behavior, in line with governance guidance distinguishing structural from prompt-layer controls \cite{imda2025}. For a partial variant, C1 additionally requires the coverage predicate to be narrowing.

\textbf{C2: behavior-trace conformance.} A candidate producing an acceptable artifact by an unacceptable route is the failure mode output comparison cannot see; C2 tests the route against the retained reference (\S\ref{sec:trace}).

\textbf{C3: production A/B, terminated by human review.} Production is the only reliable predictor of production. Under \emph{shadow}, both paths execute while only the incumbent's output takes effect; under \emph{live A/B}, arms are tested for non-inferiority against a pre-declared margin \cite{piaggio2012,kohavi2020}, restricted to in-coverage traffic to raise statistical power; under \emph{human authorization}, a reviewer examines a \emph{legible delta} and grants or withholds $H_{\text{auth}}$.

\section{Trace conformance}
\label{sec:trace}

A \emph{projection} $\pi: E \rightarrow \Sigma^{*}$ reduces an execution to a word over a checkpoint alphabet recording capability invocations, tool calls with argument-schema hashes, retrievals, guardrail outcomes, escalations, refusals, and emitted-artifact classes; free text and payload values are discarded, so a trace records what a skill did, not what it said. The spine already emits these events for provenance, so the gate is cheap. The same fact bounds its reach: a candidate calling the right tool with the wrong argument \emph{values} still conforms.

The key set $K$ is drawn from the reference's history with no human input: minimal coverage of every symbol type the reference has emitted; every trace containing an irreversible action, fired guardrail, escalation, or refusal; and every trace where the principal corrected the reference or two variants diverged. $K$ is versioned and sharpens with every subsequent correction.

Partitioning $\Sigma = \Sigma_c \uplus \Sigma_n$ into consequential and procedural symbols, the hard constraint $\pi_c(u) = \pi_c(t)$ requires exact match on consequential actions and guardrail outcomes, checked first because it is cheapest and its violation is never acceptable. The procedural remainder is compared by weighted edit distance \cite{levenshtein1966}, calibrated \emph{against the reference itself}: we replay the reference $m$ times per key trace and collect its own run-to-run divergence $D_{\text{self}}$, and a candidate passes the soft test when its divergence $D_{\text{cand}}$ from the same reference is not stochastically larger than $D_{\text{self}}$, by one-sided permutation test with FDR control across the batch \cite{benjamini1995}. The claim under test, no further from the reference's own record than the reference is from itself, is weaker than equivalence and considerably stronger than output similarity, with a tolerance the system supplies and its designers do not. Calibrating from $D_{\text{self}}$ also makes the gate's own discriminative power an observable, reportable property: where the band is wide, we mark the capability \emph{trace-opaque} and declare C2 inapplicable, instead of letting it go silently toothless, and because $D_{\text{self}}$ is estimated conditional on a candidate's claimed coverage region, opacity belongs to a region, not to a whole capability. That is the configuration habit formation wants, since the tail is precisely where opacity is expected and acceptable.

C2 returns one of three verdicts: \textsc{Pass} (hard constraint holds, soft test not rejected), \textsc{Decline} (hard constraint violated on a labeled entry, auto-declined), or \textsc{Escalate} (hard constraint holds, soft test rejected, divergence concentrated on procedurally unlabeled entries, proceeding to C3 as an improvement hypothesis under superiority framing). A binary pass/fail would reject every candidate that improves on the reference. The third verdict exists to catch those. Generation that can see $K$ will optimize to it, so $K$ is split roughly three-to-one into $K_{\text{forge}}$ and a $K_{\text{sealed}}$ portion never exposed to any generation mechanism; a candidate passing on the forge set and failing the sealed set indicts the generating mechanism rather than the individual candidate.

\section{Recovery}
\label{sec:recovery}

Recovery is a coverage withdrawal: narrowing or revoking a variant's predicate returns its traffic to the next level up, continuously, without a rebuild, and requires no authorization in the direction of less autonomy. It is triggered by a fired guard, sustained shadow divergence, a concept-drift detector on guard firing rate \cite{bifet2007,gama2004}, or a human revoking $H_{\text{auth}}$. Shadow sampling is funded as a percentage of realized savings, keeping habit formation net-positive. Borrowing outcome devaluation from behavioral neuroscience \cite{gillan2014}, we periodically devalue in simulation the objective a variant serves and check whether it stops. A healthy habit remains goal-sensitive; a \emph{compulsive} one does not, corresponding in the options formalism \cite{sutton1999} to a termination function that has lost dependence on the option's own reward.

\section{Governance}
\label{sec:governance}

Generation is unattended and safety-neutral, since it produces no effects. A human authorizes admission by policy, in advance, fixing which capability classes may host candidates, the maximum unreviewed coverage, and the C2/C3 parameters. C0--C2 then run unattended inside that envelope, which acts as a runtime shield \cite{bloem2015,alshiekh2018} authorized once and enforcing continuously. A human authorizes each promotion over a legible delta. The agent never sets its own autonomy. Article 14 of the EU AI Act requires oversight commensurate with risk, autonomy, and context, enumerating operator capabilities to understand, monitor, intervene, override, and decline \cite{fink2024}. What this architecture implements against those capabilities is delta authorization: a promotion is a discrete, evidenced, reversible act whose blast radius is bounded by its coverage predicate. Whether that satisfies the Article is a legal question this paper does not attempt to answer, and we make no compliance claim. Moving the human from catching errors to authorizing changes is a bet, and the operator-complacency literature shows it can be lost \cite{bainbridge1983,parasuraman2010}. So we instrument the reviewer: time to decision, rejection rate, and evidence-panel expansion rate, per reviewer and per capability. Sustained degradation suspends unattended admission for that capability, where degradation means approvals in seconds, evidence never opened, and no rejection in a quarter.

\section{When habit formation pays}
\label{sec:pays}

Let $c_R$ be per-execution reasoning cost, $c_A$ one-time admission cost, $c_M$ per-execution monitoring cost, $c_K$ per-period fallthrough-maintenance cost, and $N$ expected in-coverage executions before the region's distribution shifts. Promotion pays when $N c_R > c_A + N c_M + c_K$. Partial coverage improves both sides: $N$ counts head-cluster executions, the majority by construction, and $c_K$ falls because tail traffic exercises the fallthrough target at no additional cost. The inequality plausibly fails at design-partner scale for tenant-owned procedural skills with execution counts in the hundreds per year; there, auditability is the honest motivation, not cost, and it should be priced as assurance rather than disguised as savings. We state that pivot as a claim we have not tested. No experiment in this paper shows an auditor better off, faster, or more accurate on a habit-formed trace than on a reasoned one, and until one does, the auditability motivation stays an argument and is not yet a result. Reviewer time-to-decision over matched traces is the experiment that would settle it.

\subsection{Personalization erodes the economics that justify it}
\label{sec:sharing}

A coverage predicate is mined from the traffic a skill actually saw, so it records one principal's way of asking and the variant fits it. The mechanism delivers per-tenant personalization without anyone authoring a per-tenant skill, and our own data shows how sharp that fit is: a guard authored on one set of phrasings covered $0$ of $40$ paraphrases of the same questions (\S\ref{sec:evidence}). Coverage did not survive a change of wording, let alone a change of tenant.

That cuts against the split above. Shared skills are the class where the cost inequality holds comfortably, because executions aggregate across tenants, and every promotion moves work onto a variant fitted to one tenant's region. The population over which $c_A$ amortizes fragments as specialization succeeds, so the break-even that justified the first promotion is not the one facing the tenth. The contract and the reasoned anchor are the objects that stay general, since the anchor is fitted to nobody, and specialization lives below them. Whether key traces and induced guardrails can be pooled across tenants without leaking one tenant's data into another's admission decisions is the question this raises, and we do not answer it.

\section{Preliminary evidence}
\label{sec:evidence}

We instantiate the smallest slice of the architecture that is checkable without the full C0--C3 apparatus: coverage-gated dispatch and the cost inequality of \S\ref{sec:pays}. The workload is self-authored text-to-SQL, chosen for the shape the paper motivates: recurring query templates parameterized by varying literals, scored by execution against a live database, not by string match. Four reasoning arms (Claude Sonnet 5, Kimi K2 Thinking, GLM-4.6, Qwen3-235B-Thinking) generate SQL from natural-language questions across 14 templates $\times$ 3 instances, 3 repeats each (504 calls). A single deterministic candidate, one regex-gated parameterized template per shape that admits or declines each input and never guesses, is compared against each arm by a one-sided non-inferiority test (5-point margin, $\alpha=0.05$, Benjamini--Hochberg corrected across arms).

\textbf{Reproducibility is where the gap is widest.} Ask a reasoning arm the same question three times and count how often all three answers are byte-identical. Sonnet manages it on 26 of 42 questions; GLM on 13; Qwen and Kimi on 11 each. Three of the four arms reproduce their own output on a quarter to a third of the questions they are asked, and their mean pairwise edit distance across repeats runs $0.05$ to $0.23$. A habit-formed variant reproduces on everything it serves, because a script has nothing to vary. On the dispatch substrate of \S\ref{sec:dispatch}, all $456$ served inputs returned an identical route and identical parameters across five repeats, over $2{,}280$ calls with no disagreement. This is the architecture's clearest effect. It needs no interval and no noise floor, since an output either reproduces or it does not. A buyer or an auditor can check it without knowing anything about the task.

\begin{figure}[t]
\centering
\begin{tikzpicture}
\begin{axis}[
  width=\columnwidth, height=5.2cm,
  xmode=log, xmin=1, xmax=1000,
  ymin=-25, ymax=62,
  xlabel={Reuses of a habit-formed template, $N$},
  ylabel={Net token reduction},
  yticklabel={\pgfmathprintnumber{\tick}\%},
  xtick={10,100,1000}, xticklabels={10,100,1000},
  legend style={font=\scriptsize, at={(0.97,0.35)}, anchor=east,
                draw=none, fill=none},
  tick label style={font=\scriptsize},
  label style={font=\scriptsize},
  grid=major, grid style={dashed, gray!25},
]
\addplot[domain=1:1000, samples=200, thick, color=orange]
  {(1 - (888*x + 7691)/(1998*x)) * 100};
\addlegendentry{Kimi-K2 (1998 tok/req)}
\addplot[domain=1:1000, samples=200, thick, color=yellow!70!black]
  {(1 - (888*x + 7691)/(1365*x)) * 100};
\addlegendentry{Qwen3-235B (1365)}
\addplot[domain=1:1000, samples=200, thick, color=teal]
  {(1 - (888*x + 7691)/(1348*x)) * 100};
\addlegendentry{GLM-4.6 (1348)}
\addplot[domain=1:1000, samples=200, thick, color=purple!60!pink]
  {(1 - (888*x + 7691)/(1032*x)) * 100};
\addlegendentry{Sonnet (1032)}
\addplot[domain=1:1000, samples=2, dashed, gray] {0};
\end{axis}
\end{tikzpicture}
\caption{Token reduction net of the cost of forming the habit. The
one-time charge is 7{,}691 tokens per template, covering the reasoning
that decides what to script and the authoring of the coverage guard and
templates; each served request then costs 888 tokens against the
reasoning arm it replaces. Curves cross zero at 7 (Kimi), 16 (Qwen), 17
(GLM), and 53 (Sonnet) reuses, and approach the gross saving of
13.9--55.5\% thereafter. Habit formation loses against the arm that
planned least until roughly fifty reuses, which is the regime
\S\ref{sec:pays} identifies as paying for auditability rather than for
tokens.}
\label{fig:net}
\end{figure}
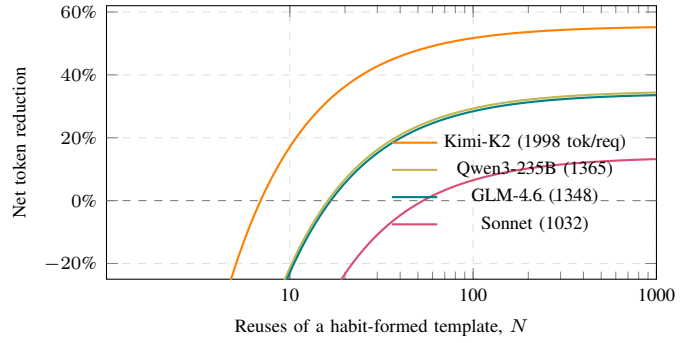

\textbf{Where the tokens go.} The same workload shows what a reasoning arm spends its tokens on. This result is secondary to replay, and its conditions matter more. Across the three arms whose reasoning tokens are metered separately from their output, reasoning accounts for $95.3$--$97.2\%$ of everything the model generates (GLM $95.3\%$, Qwen $96.2\%$, Kimi $97.2\%$), and for $62.7$--$75.0\%$ of all tokens billed, on a task whose plan is identical across all 42 questions and all three repeats. The answer itself runs to 34--44 tokens, a single SQL statement; the rest is the arm re-deriving a procedure it has already derived. We report both denominators because the cost figures below are computed over the second. Sonnet's API does not break reasoning tokens out of its output count, so we exclude it from that range instead of recording a zero: it emitted a thinking block on 4 of its 126 calls under adaptive thinking, and returned the highest accuracy ($0.865$) and the highest pass$^3$ ($0.857$) of the four arms. The arm that planned least scored best. The planning bought nothing on this workload.

The candidate is non-inferior to all four arms ($p<0.0001$ throughout), at effectively zero marginal cost and pass$^3=1$ against the arms' measured $0.76$--$0.86$. The cost inequality of \S\ref{sec:pays} therefore holds under measurement, on the one slice we can check today.

\textbf{What habit formation saves, in tokens.} Cost ratios depend on provider pricing, so we report the underlying token counts. A served request costs the reasoning arms 1032 tokens (Sonnet), 1348 (GLM), 1365 (Qwen), and 1998 (Kimi). The same request served by a habit-formed dispatch costs 888 tokens, of which 856 are input and 32 output. That is a reduction of $13.9\%$ to $55.5\%$ depending on the arm replaced, and $38.1\%$ pooled. The spread tracks planning: Sonnet thought on 4 of 126 calls and sets the floor, while the three arms that planned on almost every call set the ceiling. Almost all the saving comes from output tokens collapsing: the arms emit 99 to 1541 output tokens and the guard emits 32. The guard's input stays large because its prompt carries all ten template descriptions, which also bounds how far this scales as templates multiply. Those are gross figures. Forming the habit costs tokens of its own: the agent reasons about which part of the skill to script, then writes the coverage guard and the templates. We measure that at 76{,}910 tokens for the ten-template bundle, or 7{,}691 per template, and charge it against the savings. Net reduction after $N$ reuses of a template is $1 - (888N + 7691)/(c_R N)$, plotted in Fig.~\ref{fig:net}. Break-even arrives at 7 reuses against Kimi, 16 against Qwen, 17 against GLM, and 53 against Sonnet; at 100 reuses the net saving is $51.7\%$, $29.3\%$, $28.4\%$, and $6.5\%$ respectively. Habit formation therefore loses on tokens against the arm that planned least until a template is reused about fifty times. That is the regime \S\ref{sec:pays} identifies as paying for auditability rather than for cost.

The retrieval-routing domain saves differently and more simply: the habit-formed router issues no model call at all, because it is a local embedding lookup, so its per-decision token cost is zero by construction and not by measurement.

We instrument the admission cost $c_A$ itself and assume nothing about it: an agentic authoring loop (tool-using, same model family) derives the coverage predicate and template from two labeled examples per shape, validated against a held-out third example never shown to the model. Authoring an individual shape costs \$0.010--0.012; bundling multiple shapes into one authoring session amortizes shared context and roughly halves this to \$0.006--0.007 per shape, with every session, including all 14 shapes bundled at once, passing training and held-out validation on the first attempt. Combined with the measured $c_R$ per arm (\$0.0018--0.0043/call, recosted at standard provider pricing after an introductory rate expired), the break-even reuse count $N^{*} = c_A / (c_R - c_{\text{cand}})$ is $1.3$--$6.7$ uses depending on arm and bundle size, an order of magnitude below the design-partner-scale regime in which \S\ref{sec:pays} expects the inequality to fail.

We read this as evidence that the inequality's direction is real, and that its crossover point can be small, not as evidence that it holds generally. Every template in this pilot was authored, and every held-out example passed, on a workload chosen for structural regularity: fixed sentence skeletons with substitutable literals, unambiguous parameter boundaries, and single-template-family SQL. A held-out instance in this pilot differs from its training siblings only in the literal parameter value, never in phrasing, so the clean result above does not by itself distinguish genuine generalization from matching a memorized sentence skeleton. We ran a follow-up to separate the two.

\textbf{Complexity and phrasing are different axes, and only one of them breaks the guard.} Eighteen harder shapes (54 instances), authored via the same loop, reach the same held-out pass rate as the easy set (18/18 against 14/14) at $1.8\times$ the per-template authoring cost (\$0.0100 against \$0.0057). The harder set covers anti-joins via \texttt{NOT EXISTS}, correlated category-averages, top-$N$ by revenue, percent-of-total, conditional pivots, nth-highest via \texttt{OFFSET}, \texttt{HAVING} on aggregates, double-\texttt{EXISTS} with status, window running-totals, \texttt{RANK}, and above-overall-average. SQL complexity is therefore a cost gradient for this authoring loop and not an accuracy barrier, likely because the loop tests its own candidate against real data before submitting. Phrasing is where the guard breaks. Evaluated against 40 questions asking the same thing in different words (10 templates $\times$ 4 distinct phrasings), never shown during authoring, the hand-authored candidate and a freshly LLM-authored one both fall to \emph{zero} coverage, 0/40, so the collapse is not an artifact of hand-authoring. Inspecting the authored patterns shows why: each regex matches a literal phrase fragment copied almost verbatim from its training sentence (an exact match on the phrase ``How many orders has \dots\ placed?'' or on the substring ``more than $N$ orders''), so a coverage predicate authored this way functions as a compressed record of the exact sentences it was shown, not a semantic classifier, and it declines everything else by construction. The guard is behaving as designed: it prefers precision to recall, so an unrecognized paraphrase falls through to reasoning instead of misfiring on it. But it means the achievable coverage of literal-regex authoring on natural, non-templated traffic is likely far below what this pilot's 100\% figures suggest, and closing that gap is a guard-\emph{design} question (semantic or embedding-based matching, or an LLM-authored classifier in place of a literal pattern) rather than a task-complexity question. A third, smaller probe halving training examples from two to one degrades held-out pass rate on a three-shape subset to 2/3 ($n=3$, not conclusive alone, but consistent with the same story: less-constrained authoring generalizes less).

\textbf{The break is repairable by guard redesign alone.} We ran the concrete next experiment named above: replace the literal-regex coverage predicate with a cheap classifier call (Claude Haiku~4.5) at dispatch time, authored by the identical tool-using loop but validated against the live classifier instead of a byte-level regex match. The classifier sees only template descriptions and parameter names, never the training sentences or gold SQL, so a match on a held-out paraphrase cannot be an artifact of literal leakage. On the same 40 held-out paraphrases that collapsed the regex guard to zero, the semantic guard recovers almost all the lost coverage: 39/40 covered ($97.5\%$) at $97.4\%$ accuracy among covered, against 0/40, plus 10/10 on a held-out canonical set. One paraphrase went uncovered. With ten templates competing for a single classifier call, the guard declined instead of guessing, as the architecture asks of it. Unlike the regex guard, which had no marginal cost, the semantic guard makes a classifier call costing \$0.00102 per dispatch at this bundle size (up from \$0.000505 at five templates, since the prompt carries every competing template description); recomputing $N^{*}$ with this nonzero $c_{\text{cand}}$ still clears break-even at $5.4$--$24.1$ uses per template across the four reasoning arms. Break-even sits furthest out against the cheapest arm, whose $c_R - c_{\text{cand}}$ gap is smallest. Closing the paraphrase gap was, as hypothesized above, a guard-\emph{design} question rather than a ceiling on coverage-gated dispatch itself: the candidate body (the SQL template and its parameters) was untouched between the two guard designs; only the predicate changed.

\textbf{What the guard lets through.} Coverage is reported above as a benefit, and the coverage that is won is also exposure. A declined input is safe: it falls through to reasoning and costs only tokens. An input the guard \emph{admits} and serves wrongly is the failure \S\ref{sec:lineage} names as the architecture's worst, because no reasoning step ran in which the mismatch could have surfaced. We therefore report the guard's \emph{false-admission rate}, meaning inputs it claimed and answered incorrectly, as a first-class quantity alongside its coverage.

On the 40 held-out paraphrases above, the semantic guard's $97.4\%$ accuracy-among-covered is one wrong answer in 39: a false-admission rate of $2.6\%$ $[0.5\%, 13.2\%]$. The single case is instructive. Asked ``In total, how many product-category support tickets have been resolved?'' the guard selected the correct template and extracted the category as \texttt{product-category} rather than \texttt{product}, producing well-formed SQL that executed against the live database and returned $0$ where the answer was $19$. The template was right, the route was right, and the served answer was confidently and silently wrong.

To find where this behavior concentrates, we scored the same guard on 80 further items built to sit near its boundary: questions between two similar templates, questions whose parameter value collides with a category name or template keyword, questions a template plausibly matches but cannot express, and questions out of scope entirely. Each carries either a gold template and validated SQL or an explicit \emph{null} meaning no template should serve it. This is a robustness stress set and not an adversarial evaluation: nothing is obfuscated, no adversary is modeled, and the items were authored to probe the boundary rather than to defeat the guard.

The guard admitted $50$ of $80$ and declined $90.6\%$ $[75.8\%, 96.8\%]$ of the items no template should serve, including all 13 that were out of scope. That is the precision-over-recall asymmetry behaving as designed. Among the admitted, the false-admission rate is $26.0\%$ $[15.9\%, 39.6\%]$. It is not distributed the way the architecture's own account of the hazard predicts. Boundary cases between similar templates, the failure mode \S\ref{sec:limits} anticipates, produce $4.2\%$; parameter-value collisions produce $39.1\%$. The guard is good at deciding \emph{which} procedure applies and much weaker at reading \emph{what} to give it: $9$ of the $13$ false admissions are correct-template, wrong-parameter, and three of those differ from the gold value only in capitalization.

\textbf{The central gate cannot see most of them.} Crosswalking each false admission against C2 (\S\ref{sec:trace}), we find $11$ of $13$ invisible to it, for two compounding structural reasons that no amount of tuning reaches. First, $\Sigma_c$ is empty for a read-only query, so the hard constraint never fires in this domain at all. Detection therefore rests entirely on the soft test over $\Sigma_n$, which records argument \emph{schemas} and not argument \emph{values}. A wrong-parameter serve emits a $\Sigma_n$ word identical to the reference's, so $d_{\text{cand}} = 0$; a deterministic template has $d_{\text{self}} = 0$; and the permutation test is comparing a distribution against itself. No choice of $\alpha$ and no number of replays $m$ can reject. This is not a gate that usually misses these cases. It is a gate for which they are undetectable. Discarding payload values is what keeps the projection of \S\ref{sec:trace} cheap enough to run on every candidate; these cases are the price of that choice, and we measure it. Lifting declared argument predicates into $\Sigma$ (cheap deterministic invariants of the same character as guardrails) is the mitigation this result argues for most sharply.

Two limits on that number. The stress set is ours and was written to find the boundary, so $26\%$ is a rate on adversarially-shaped-but-benign inputs, not an estimate of production traffic; the $2.6\%$ on naturally-phrased paraphrases is the better guide to the latter. And $n$ is small enough that the interval spans a factor of two.

\textbf{A second domain: the mechanism travels.} The evidence above lives on one workload, and its own follow-ups showed that phrasing breaks the guard and complexity does not. The obvious reviewer question is whether that is a property of habit formation or of this pilot's construction, so we replicated the coverage-guard argument at a different point on the phrasing/complexity surface: \emph{retrieval routing}. A router decides, per question, whether to retrieve from a knowledge base or answer directly; the skill is the route-then-retrieve procedure. Unlike the SQL workload, this one is not ours: the questions and their supporting corpus are seeded from HotpotQA \cite{yang2018hotpotqa}, and the loop runs on a third-party agent platform we did not write. What remains ours is the paraphrase expansion and the retrieve-versus-answer labels, so the decision boundary is still one we defined even though the task is not. Scoring is on a frozen 288-question test pool, split into intents seen during mining and intents never seen. The guard is mined \emph{autonomously}: a misroute is inferred from the router's own trace plus a grader's verdict on the resulting answer, never from gold routes, which are held back and used only to score the signal after the fact (measured precision $0.70$--$0.83$). Generation is held fixed, and the grader is a different model family from the generator, so answer quality reflects routing rather than generation luck or self-enhancement.

\begin{figure*}[t]
\centering
\begin{tikzpicture}
\begin{groupplot}[
  group style={group size=3 by 1, horizontal sep=26pt,
               ylabels at=edge left, yticklabels at=edge left},
  width=0.36\textwidth, height=0.285\textwidth,
  xmin=-12, xmax=212, ymin=0.5, ymax=0.96,
  xtick={0,50,100,200}, ytick={0.5,0.6,0.7,0.8,0.9},
  yticklabel style={/pgf/number format/fixed,
                    /pgf/number format/precision=1},
  xmajorgrids, ymajorgrids, grid style={black!12},
  axis line style={black!45}, tick style={black!45},
  xlabel={Mined traces},
  title style={font=\small}, label style={font=\small},
  tick label style={font=\footnotesize},
]
\nextgroupplot[title={GLM-4.6}, ylabel={Routing accuracy}]
\addplot[draw=none, fill=okblue, fill opacity=0.22, forget plot]
  coordinates {(0,0.6554) (25,0.8445) (50,0.8644) (100,0.8569)
    (200,0.9206) (200,0.6928) (100,0.7357) (50,0.6981) (25,0.6486)
    (0,0.6455)} -- cycle;
\addplot[black!55, dashed, forget plot]
  coordinates {(-12,0.6505) (212,0.6505)};
\addplot[okblue, thick, mark=*, mark size=1.6pt]
  coordinates {(0,0.6505) (25,0.7465) (50,0.7812) (100,0.7963)
    (200,0.8067)};
\nextgroupplot[title={Kimi-K2}]
\addplot[draw=none, fill=okverm, fill opacity=0.22, forget plot]
  coordinates {(0,0.6840) (25,0.8803) (50,0.8572) (100,0.8809)
    (200,0.8673) (200,0.7289) (100,0.7201) (50,0.7192) (25,0.7165)
    (0,0.6840)} -- cycle;
\addplot[black!55, dashed, forget plot]
  coordinates {(-12,0.6840) (212,0.6840)};
\addplot[okverm, thick, mark=*, mark size=1.6pt]
  coordinates {(0,0.6840) (25,0.7984) (50,0.7882) (100,0.8005)
    (200,0.7981)};
\nextgroupplot[title={DeepSeek-v3.2}]
\addplot[draw=none, fill=okgreen, fill opacity=0.22, forget plot]
  coordinates {(0,0.5938) (25,0.8566) (50,0.8588) (100,0.8835)
    (200,0.9288) (200,0.6591) (100,0.7368) (50,0.7176) (25,0.4976)
    (0,0.5938)} -- cycle;
\addplot[black!55, dashed, forget plot]
  coordinates {(-12,0.5938) (212,0.5938)};
\addplot[okgreen, thick, mark=*, mark size=1.6pt]
  coordinates {(0,0.5938) (25,0.6771) (50,0.7882) (100,0.8102)
    (200,0.7940)};
\end{groupplot}
\end{tikzpicture}
\caption{Habit formation on retrieval routing: routing accuracy against
mined traces, three routers, three seeds each. The line is the seed
mean, the band the 95\% confidence interval, and the dashed rule each
router's cold start. Every router gains from mining and then holds:
pooled over all nine runs the cold-start-to-$t{=}200$ gain is $+0.157$
$[+0.116, +0.198]$, while the pre-specified $t{=}100 \rightarrow
t{=}200$ contrast is $-0.003$ $[-0.028, +0.022]$. Band widths are
comparable to the $0.118$ same-configuration replication spread of
\S\ref{sec:evidence}, which is why nine runs rather than one stand
behind each interval. Non-habit-forming baselines are flat at every
checkpoint and are omitted.}
\label{fig:dose}
\end{figure*}
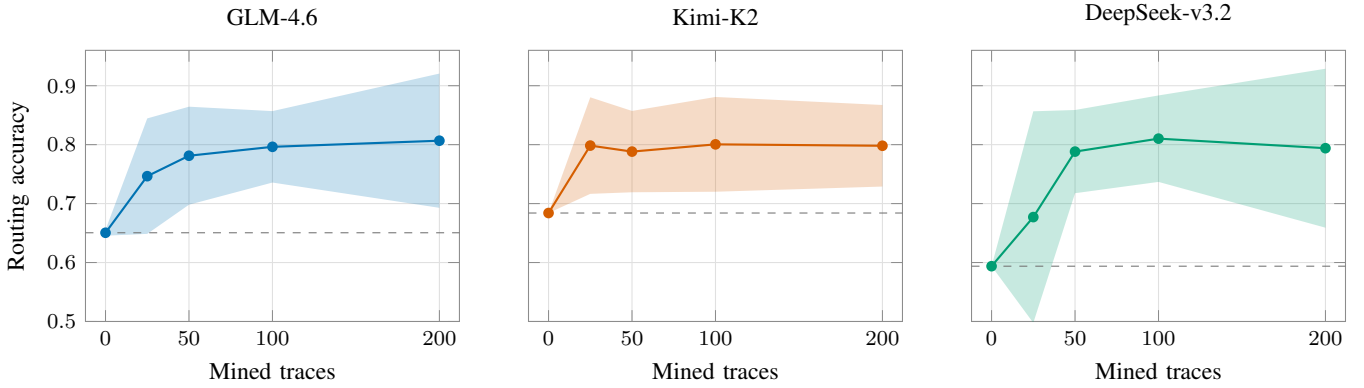

Two non-habit-forming baselines anchor the comparison: a minimal static classifier and one hardened once by a stronger teacher model. Both are floors, not strong competitors. A dose-response design needs floors. They are flat at every checkpoint by construction, so a rising curve is the treatment and not drift. Against both, a guard that folds mined corrections into its own few-shot block improves every router it was run on, from a cold start. The result that does not follow from the SQL domain is the generalization split, and the grid behind it is ragged in a way worth stating before the numbers. Three routers were swept to 100 mined traces: GLM-4.6 and DeepSeek-v3.2 at three seeds each, and Kimi-K2 at one seed before it was dropped from the grid for provider latency. Only the first two continued to 200. We therefore report the wider pool at the checkpoint it reaches and the narrower one at the pre-specified endpoint, rather than pooling them into a single figure that neither supports.

At 100 traces, across three routers and seven runs, the gain from cold start is $+0.157$ $[+0.101, +0.214]$ in routing accuracy overall and $+0.157$ $[+0.086, +0.227]$ on intents never seen during mining. At the pre-specified endpoint of 200 traces, across two routers and six runs, it is $+0.145$ $[+0.113, +0.178]$ overall and $+0.132$ $[+0.082, +0.182]$ unseen. Intervals are over runs rather than over questions, since the 288 test questions are re-scored by every run and are not independent observations of the effect. Mining one set of intents improved routing on a disjoint set, at both checkpoints and on every router we ran. That is generalization and not memorization, and the SQL workload could not show it.

The two checkpoints differ by about two points, and the direction matters more than the gap. On the six runs that reach both, the same contrast reads $+0.168$ $[+0.106, +0.230]$ at 100 traces and $+0.145$ at 200: the curve rises to a peak near 100 and gives a little back. We name the endpoint as the endpoint and the peak as the peak, because the maximum over checkpoints is at least the final value by construction and quoting it as an endpoint would inflate the headline while misdescribing the shape. Mining bought most of its value early, and the last hundred traces bought nothing measurable. The near-zero pre-specified $t{=}100 \rightarrow t{=}200$ contrast below is the same finding, not a competing one.

\textbf{How large an effect this workload can resolve.} Dose-response curves invite reading structure into them, so we measured the floor first. Independent replications of an identical configuration at an identical seed diverge by up to $0.118$ in routing accuracy here. Temperature-0 decoding is not run-to-run deterministic \cite{he2025nondeterminism}, and a mining loop compounds that: a differing answer draws a differing grade, which mines a differing correction, which changes every subsequent routing decision. Any feature of a single curve smaller than that band is therefore unfalsifiable. That is why every interval above is over runs rather than over the questions they re-score: six runs for the endpoint contrast, seven for the checkpoint at 100. Two consequences follow for anyone measuring a habit-forming system. Replication spread should be published beside the curve, as the floor a claim has to clear. And the contrast should be named before the run: peak-to-final is a natural thing to reach for and is biased, since the maximum over checkpoints is at least the final value by construction, so that statistic is non-positive for every run whatever the data show. Pooled over all nine habit-forming runs, the six refine runs and the three memory runs, it reads $-0.015$ where the pre-specified $t{=}100 \rightarrow t{=}200$ contrast is $-0.003$ $[-0.028, +0.022]$. We report the pre-specified contrast, and find no reliable non-monotonicity at the doses we tested. Whether habit formation degrades at larger doses remains open; our data bounds the effect near zero at $t{=}200$ rather than showing that no such effect exists.

\textbf{What a capability's growth costs its guard.} \S\ref{sec:limits} names ambiguity among competing templates as an untested hazard of scale. We tested it by sweeping bundle size from 5 to 30 templates over a fixed question set, so that only the number of procedures competing for each decision varies, with distractor subsets nested within a seed and bundle order shuffled.

The guard's prompt carries every competing description, so its cost is linear in bundle size: $\$0.000548$ per dispatch at 5 templates rising to $\$0.001969$ at 30, with input growing from 403 to 1800 tokens. A least-squares fit gives $c_{\text{cand}}(N) = \$0.000263 + \$0.0000564N$ with $R^2 = 0.999$.

This is not the claim that more templates make a capability more expensive, and the design is why. Coverage here is flat by construction: the evaluation set is fixed, five core templates serve it at every bundle size, and the templates added to reach $N$ are distractors that claim no traffic. The sweep therefore isolates what bundle growth costs while holding constant what it buys. That is the worst case, not the operating one. In deployment a template is added \emph{because} it claims traffic, and the system cost per request is $c_{\text{cand}}(N) + (1 - \text{cov}(N))\,c_R$, in which the second term falls as the first rises.

What the linearity gives us is the exchange rate between them. The marginal template adds $\$0.0000564$ to \emph{every} dispatch the capability serves, including the ones it does not claim, and repays that only on the share it does. A new template is therefore net positive exactly when the traffic share $s$ it claims satisfies $s > \$0.0000564 / c_R$: between $1.3\%$ and $3.2\%$ across our four reasoning arms, $3.2\%$ against the cheapest. Below that floor a template loses money \emph{even when it is perfectly accurate}, because accuracy is not the quantity being traded.

That threshold is the utility problem \cite{minton1990} made concrete, and it is the minimum coverage floor \S\ref{sec:limits} calls for. We derived it from measurement instead of picking it. It also bounds the earlier break-even range. The 7-to-53 figure holds coverage fixed, so it describes a capability that accumulated templates without accumulating traffic. That is the failure the floor exists to prevent, not an operating point we expect. A lineage whose templates each clear the floor gets cheaper as it grows; one that admits narrow variants below it gets more expensive with every promotion, and no gate in \S\ref{sec:gates} currently measures the quantity that decides which of those is happening.

The safety result refutes what we expected. Ambiguity among competing templates did not appear: across 2618 dispatches there is not one wrong-template admission, and in-scope traffic is served correctly at every size (0 false admissions in 1870 covered in-scope dispatches, Wilson upper bound $0.21\%$, coverage $100\%$ throughout). What degrades is the decline. All 29 false admissions are out-of-scope questions routed to a template that cannot serve them, and every one required two conditions at once: a semantically adjacent template present in the bundle, \emph{and} a bundle of at least 20. Adjacency without size produced none in 66 attempts; size without adjacency produced none in 154. Together they produce $18.8\%$ $[13.4\%, 25.7\%]$.

The distinction is worth keeping. Scale does not make this guard worse at serving work its capability covers; it makes it worse at refusing work its capability does not. Template count alone is not the hazard, and neither is adjacency alone. The hazard is what a bundle \emph{contains} as it grows, not how large it grows. That is a property of how a lineage accumulates.

\textbf{Where the routing guard's errors go instead.} The same accounting on the routing domain returns a result we did not expect, and we report it as it stands. That guard has no abstain path: it emits a route for every input and defers nothing, so its coverage is $1.0$ by construction and its false-admission rate is numerically its error rate, inheriting the $0.118$ replication floor exactly. Habit formation lowers it, from $0.382$ to $0.237$ between cold start and 200 mined traces, a move of $-0.145$ that clears the floor and is down in all six runs.

The total is the wrong thing to watch. Decomposed by direction, the entire improvement comes from the cheap error, retrieving when answering directly would have served, which falls $-0.216$. The severe error, answering directly and unsourced where retrieval was required, moves from $0.000$ to $0.071$: from structurally impossible to present. The cold guard retrieves so readily that it cannot commit the severe error at all. Habit formation is what makes it capable of one. \textbf{That $+0.071$ sits below the $0.118$ floor, and we do not claim it as an effect.} What survives the floor is its direction, unanimous across all six runs, and its mechanism, which is not in doubt: a guard that always retrieves has no way to fail by not retrieving. We report it for two reasons. A single aggregate number hides this case, where the headline metric improves while the worst failure class goes from absent to present. And the teacher-hardened baseline already carries a false-direct rate of $0.046$ with no mining at all, so trace mining does not introduce this failure class to the architecture.

\textbf{An admission gate can deadlock the learner it filters.} One result is architectural and bears directly on the C0--C3 design. Our routing runs included an embedding-memory guard whose dose-response curve was perfectly flat. That flatness came from the admission gate, not from embedding memory. Its admission log records every batch rejected, and its memory was empty at the end: a cold memory-based guard routes everything one way and so scores like a majority-class predictor, a candidate built from a handful of corrections scores slightly below that, and the non-inferiority gate refused it, correctly on its own terms. The gate did not slow habit formation; it prevented it, and produced a curve that reads as ``this approach does not work.'' Non-inferiority admission is hardest on a learner when the learner has the least evidence, and an incumbent that is uniformly mediocre is a deceptively hard target.

\textbf{Autonomously mined corrections cannot, alone, train a guard with no prior.} Banking the rejected evidence across attempts left the deadlock in place, and the reason proved more useful than the fix. Mined corrections are the current policy's \emph{errors}, so their labels track whichever direction that policy over-commits: a cold guard routing everything directly mined 69 corrections, all of one class; a fallback that over-retrieved mined 55, of which 50 were the other. Neither is a class-balanced record, and no amount of mining makes one. A majority-vote memory over single-class evidence is itself a constant classifier, just the opposite constant. This explains the split between our two habit-forming guards. The classifier guard improves because it already routes both ways on its own and a one-sided nudge is all it needs; the memory guard cannot, because the mined record \emph{is} its decision function. Mining the non-errors as well, since a trace whose signal did not fire is autonomous evidence its route was acceptable, restores class balance on every seed and recovers $+0.131$ from cold start, reaching parity with the teacher-hardened static baseline while making no model call at inference. But it moves the unseen-intent split by $-0.023$, to $+0.023$: pure memory buys \emph{coverage}, not generalization, which is consistent with the coverage semantics of \S\ref{sec:lineage} rather than a contradiction of them.

\subsection{Which half of dispatch needs a model}
\label{sec:dispatch}

The guard above does two jobs in one call: it picks a template, and it fills that template's parameters. We separated them, on a substrate of templates mined from traces at catalog sizes $5$ and $20$ and reuse depths $5$, $20$, and $50$, scored on a frozen $76$-item set that includes questions no template should serve.

Routing is solved and extraction is not. The one-call guard routes correctly on $456$ of $456$ decisions across all six cells, and $51$ of its $51$ false admissions are wrong-parameter. Zero are wrong-template. The largest single fault is arithmetic: where a mined template compares \texttt{order\_date < \{end\_date\}}, the guard supplies the last day of the month instead of the first day of the next, and it gets $0$ of the $51$ exclusive-window slots right. Any extractor told the slot's comparison operator gets all $51$ right.

Splitting the call improves both numbers. One small call that already knows the route, carrying only that template's slots, reaches $0.952$ end-to-end against the joint call's $0.888$, at $43\%$ of the cost. A deterministic extractor that resolves each slot against the live database schema and fails the slot instead of guessing reaches $0.925$ end-to-end at zero marginal cost, with extraction correct on $184$ of the $200$ items it commits to. Against the joint call it recovers 19 wrong answers, converts 16 more into declines, and costs 2 the joint call had right. A deployment has to price that trade. Two arms we expected to help did not. A cascade from the deterministic extractor to the model is bit-identical to the deterministic extractor alone in all six cells and costs strictly more. Constrained decoding is worse than leaving it off: forced to choose inside a $38$-name customer domain, the decoder substitutes one real customer for another and cannot fail a slot, so it converts declines into confident errors.

Routing resists the same treatment. A classifier over local embeddings, making no model call, ranks the correct template first on $53$ of $53$ in-catalog items, yet its routing accuracy is $0.776$ against the model guard's $1.000$, because its abstention threshold discards $30\%$ of argmaxes that were already right in order to decline $22$ of $23$ questions it should refuse. Lexical matching cannot abstain at all: its top-$1$ score is higher on out-of-catalog questions than in-catalog ones, because those questions are longer. The deficit is calibration. Ranking is already perfect, and fitting the threshold on the evaluation set, an oracle bound we report only to size the gap, recovers about a third of it.

Two things follow for the architecture. The expensive half of dispatch is the half earning its money, and the cheap half is the one producing the errors, which inverts what a cost-reduction reading of this paper predicts. And the guard's prompt carries every competing description, so at $20$ templates it spends $2360$ tokens per dispatch, more than the $1032$ to $1998$ the reasoning arms spend on the request it replaces. Beyond a catalog of that size the current guard design stops paying for itself on tokens alone, whatever its coverage.

\section{Limitations}
\label{sec:limits}

\textbf{The coverage guard is ungated.} Every other component faces C0--C3, while the guard is trained alongside the variant and thereafter decides unsupervised whether reasoning is skipped. Its failure mode is quiet: a false admission means no reasoning ran, so nothing was positioned to notice. Boundary-weighted shadow sampling, a held-out conformance test for the guard itself, and margin-gated escalation each address it, and none is evaluated here. \S\ref{sec:evidence} now measures the rate rather than only naming the hazard.

\textbf{One of our two guards cannot decline at all.} The architecture requires a guard to have somewhere to defer to, and the SQL guard does: it returns no template and the request falls through. The routing guard does not. It emits a route for every input, defaults to one of them when its own response fails to parse, and computes a confidence that no caller reads, so its effective coverage is $1.0$ by construction and no input in $30{,}240$ scored decisions was ever deferred. The design called for escalation on ambiguous questions, and it was not built. We report this rather than narrowing the claim to the domain that behaves. One of the two domains in \S\ref{sec:evidence} shows the architecture's cost savings without the safety property those savings are meant to buy. And a guard that cannot abstain is the configuration in which every error it makes is a false admission.

\textbf{Narrow claims tile into wide ones.} A candidate claiming one percent of traffic clears every gate on weak evidence, and a hundred such candidates tile the space one full-coverage variant would have had to earn. Admission evidence is evaluated per variant while risk accrues per chain. A minimum coverage floor and chain-level evidence accounting both follow.

\textbf{Anchor warmth falls with lineage depth.} Traffic reaches the first variant that admits it, so intermediate variants absorb most declined work and the anchor sees only the deepest tail. Our reversion-safety argument holds for shallow chains and weakens as chains grow.

\textbf{Conformance filters rather than certifies.} A candidate can match every key trace while producing worse work, and assurance is measured against a baseline nobody audited. Generated candidates also carry the risks any generated code carries \cite{pearce2022}.

\textbf{Four framework quantities remain unmeasured}: achievable coverage per capability, trace-opaque fraction within coverage regions, forge yield through C0--C3, and realized $c_K$ across a model deprecation cycle. \S\ref{sec:evidence} reports two workloads and a break-even, which speaks to the cost inequality and to coverage-gated dispatch, and leaves the lineage architecture at production scale untested.

\section{Future work}
\label{sec:future}

The cost of forming a habit is forming a bad one, and we can now say where the architecture fails to prevent that.

\textbf{No gate reaches the guard.} C1 is a static subset check and does not evaluate behavior. C2 is structurally blind to the dominant fault: $\Sigma_c$ is empty for a read-only query, so the hard constraint never fires, and $\Sigma_n$ records argument schemas, so a wrong-parameter serve emits a word identical to the reference's and the soft test compares a distribution against itself at any $\alpha$ and any replay depth. That leaves C3. A non-inferiority test at a five-point margin certifies that a candidate is no worse by more than five points, so a defect at the $2.6\%$ rate we measured on natural paraphrases clears it with room to spare. The same test is also too strict in the opposite regime: in the routing domain it refused a cold memory-based learner outright and produced a flat curve that reads as a failed method. Non-inferiority admission is most forgiving when a defect is small and silent, and least forgiving when a learner has the least evidence. We would rather have a gate whose blocking decisions are reported as precision and recall than one whose blocking rate goes unreported, which reads the same as a wall.

\textbf{Admission is per candidate while risk accrues per catalog.} A template that passes every gate and runs correctly becomes unsafe when a sibling is admitted beside it. Sweeping catalog size from $5$ to $30$, false admissions appear only when a semantically adjacent template is present \emph{and} the catalog exceeds twenty: adjacency alone produced none in $66$ attempts, size alone none in $154$, and the two together $18.8\%$. Nothing re-runs admission on an incumbent when the catalog around it changes. This argues for a re-admission trigger on catalog change, and for evidence accounted at the chain.

\textbf{Production has no gold answer.} We caught the parameter faults by executing candidate SQL against a known result. A live A/B has no such oracle, and a wrong parameter returns a plausible number rather than an error. Shadow sampling is the mechanism that fits, since it compares candidate against reference and needs no rubric, but it is funded as a share of realized savings and nobody has measured what detection rate that buys at a $2.6\%$ defect. That measurement, and a legible delta that shows a reviewer the divergent cases and not only the aggregate, are prerequisites for trusting the human step.

\textbf{The discovery gate measures fit, not generalization.} Discovery success falls as evidence grows, from $1.000$ at one recorded execution to $0.900$ at fifty, because the gate asks whether a mined template reproduces the traces it was mined from. At one trace that question is vacuous. The template that resists discovery at every depth does so because its own traces contradict each other, and a held-out test would report that as the useful fact it is. Splitting mined evidence into a fitting and a held-out portion, as C2 already does for its key set, is the obvious repair and we have not run it.

\section{Conclusion}

Agents disagree with themselves on $38\%$ to $74\%$ of repeated tasks, and spend $95.3\%$ to $97.2\%$ of what they generate re-deriving plans they already hold. Habit formation addresses both. A habit-formed variant reproduced on all $456$ dispatches we repeated, where the arms it replaced managed 11 to 26 of 42, and it was non-inferior to every one of them. It cut tokens by $14\%$ to $56\%$ per served request, turned net positive after 7 to 53 reuses, and removed the router call outright on a second workload.

The price is a guard that claims work it should have deferred, on $2.6\%$ of natural traffic and $26\%$ of inputs near its boundary. Those errors are silent, because no reasoning step ran in which the mismatch could surface, and they are exactly reproducible: a bad habit is as reliable as a good one. Our central gate cannot see most of them, since it compares argument schemas and these are argument values.

Two findings apply to any system that caches procedures, not only to ours. Separating which procedure to run from what to give it raised end-to-end accuracy from $0.888$ to $0.952$ at $43\%$ of the cost, so the guard should be two mechanisms and not one. And risk accrues to a catalog while evidence is gathered per candidate: a template that passed every gate stops being safe when a sibling is admitted beside it.

Habit formation can therefore produce a bad habit as readily as a good one. A production deployment would need mechanisms to detect that a habit has gone wrong and to correct it, and designing and measuring those mechanisms is future research.


\end{document}